\documentclass[journal]{IEEEtran}
\usepackage{cite}
\usepackage[dvipsnames]{xcolor}
\usepackage{graphicx}
\usepackage{amsmath}
\usepackage{amssymb}
\usepackage{algorithmic}
\usepackage{array}
\usepackage{stfloats}
\usepackage{url}
\usepackage[caption=false, font=footnotesize]{subfig}
\usepackage{tikz}
\usepackage{comment}
\usepackage{pgfplots}
\usepackage{multicol}
\usepackage{booktabs}
\usepackage{tabularx}
\usepackage[normalem]{ulem}

\usepackage{framed}
\DeclareMathOperator{\mean}{mean}

\pgfplotsset{compat=1.13}
\usetikzlibrary{positioning,spy,backgrounds,calc,arrows}
\pgfdeclarelayer{background}
\pgfdeclarelayer{foreground}
\pgfsetlayers{background,main,foreground}

\graphicspath{{./images/}}

\begin{document}
\title{MoDL: Model Based Deep Learning Architecture for Inverse Problems}

\author{Hemant~K.~Aggarwal,~\IEEEmembership{Member,~IEEE}, Merry~P.~Mani,
        and Mathews~Jacob,~\IEEEmembership{Senior~Member,~IEEE}
        \thanks{This work is supported by NIH 1R01EB019961-01A1 and ONR-N000141310202.}%
}

\maketitle
\begin{abstract}
We introduce a model-based image reconstruction framework with a convolution neural network (CNN) based regularization prior. The proposed formulation provides a systematic approach for deriving deep architectures for inverse problems with the arbitrary structure. Since the forward model is explicitly accounted for, a smaller network with fewer parameters is sufficient to capture the image information compared to direct inversion approaches, thus reducing the demand for training data and training time. Since we rely on end-to-end training with weight sharing across iterations, the CNN weights are customized to the forward model, thus offering improved performance over approaches that rely on pre-trained denoisers. Our experiments show that the decoupling of the number of iterations from the network complexity offered by this approach provides benefits including lower demand for training data, reduced risk of overfitting, and implementations with significantly reduced memory footprint. We propose to enforce data-consistency by using numerical optimization blocks such as conjugate gradients algorithm within the network; this approach offers faster convergence per iteration, compared to methods that rely on proximal gradients steps to enforce data consistency. Our experiments show that the faster convergence translates to improved performance, primarily when the available GPU memory restricts the number of iterations.

\end{abstract}

\IEEEpeerreviewmaketitle

\section{Introduction} 
\IEEEPARstart{M}{odel}-based recovery of images from noisy and sparse
multichannel measurements is now a mature area with success in several
application areas such as MRI~\cite{fessler2010magazine}, CT~\cite{fessler2002CT},
PET~\cite{petUnser}, microscopy~\cite{unsreMicroscopyModelBased}. These schemes rely on a numerical model of the measurement system, often termed as the forward model. Image recovery is then posed as an optimization problem, where the objective is to improve the consistency between the true data and the measurements obtained from the image using the forward model. Since the recovery from few measurements is an ill-posed problem, the general approach is to modify the objective function using priors that penalize solutions that fall outside the class of natural images. Carefully engineered priors including total variation~\cite{shiqianma2008}, patch-based non-local methods, low-rank penalties~\cite{Lingala2011,lingala2012blind}, as well as priors learned from exemplary data~\cite{ravishankar2015tsp} or the measurements themselves \cite{bcs2013MathewsJacob} are widely used. 

Recently, several researchers have proposed to exploit the power of
deep convolutional neural networks (CNN) in image recovery. Some of these schemes
customize existing CNN architectures (e.g., UNET~\cite{ronneberger2015unet} \& ResNet~\cite{heCVPR2016residualLearning}) to image recovery tasks \cite{wangCTtmi2017,wangDeepImagingIEEEaccess,zhang2017dncnn}. These schemes rely on a single framework to invert the forward model and to exploit the extensive redundancy in the images. An alternative for this direct inversion approach is iterative algorithms that alternate between data-consistency enforcement and pre-trained CNN denoisers  \cite{zhang2017magazine,Zhang2017plugplaycnn,oneNetwork2017iccv}. These schemes train the CNN denoisers to denoise Gaussian noise corrupted images; since the noise and alias terms decay with iterations, different CNNs pre-trained with different noise levels are used at different iterations \cite{zhang2017magazine,Zhang2017plugplaycnn,oneNetwork2017iccv}. End-to-end training schemes \cite{schlemper2017cascadeRueckert,
  hammernik} rely on similar architecture as pre-trained models, but the CNN parameters at different iterations are trained such that the resulting deep network recovers the images from undersampled measurements. Similar recursive neural network architectures inspired by proximal gradient algorithms \cite{Mardani2018cvpr,hammernik,putzky,lista},  which alternate between CNN blocks and steepest descent steps, have also been introduced by several researchers. 

The main focus of this work is to introduce a systematic approach for the development of deep architectures for arbitrary inverse problems. The proposed framework, termed as \textit{MOdel-based reconstruction using Deep Learned priors} (MoDL), merges the power of model-based reconstruction schemes with deep learning. We use a variational framework involving a data-consistency term and a learned CNN  to capture the image redundancy. We use an alternating recursive algorithm, which when unrolled yields a deep network. The  network consists of interleaved CNN blocks, which captures the information about the  image set, and data consistency blocks that encourages consistency with the measurements. The data consistency block involves a quadratic sub-problem, which has analytical solutions for simpler problems such as single channel MRI recovery \cite{schlemper2017cascadeRueckert}. When the forward model is more complex (e.g. multichannel MRI), we propose to solve the quadratic sub-problem using conjugate gradient (CG) optimization. The use of numerical optimization CG blocks within the deep network is not reported, to the best of our knowledge. In addition to accounting for the complex forward model, this approach can also facilitate hybrid strategies that incorporate other image regularization terms. For example, we have demonstrated the utility of combining patient specific STORM priors \cite{storm} along with population generalizable deep learned priors in \cite{modlStormICASSP2018}. Since the forward model and its adjoint are part of the network, a low-complexity plug-and-play CNN with a significantly lower number of parameters is sufficient to learn the image set, compared to some existing CNN methods that do not have data consistency term.  We propose to train the framework assuming different forward models (e.g. sampling patterns), which enables the learning of a network that can be re-used in a variety of sampling conditions. 

We now briefly describe the differences between  the proposed framework and the current methods as described above. The proposed framework is very different from direct inversion schemes \cite{wangCTtmi2017,wangDeepImagingIEEEaccess,zhang2017dncnn} that cannot guarantee data consistency.  While the ability of these methods to learn the inverse is remarkable, they have some practical drawbacks. Since the receptive field of the CNN has to match the support of the point spread function, large networks with many parameters (e.g., UNET) are often needed in applications including tomography and Fourier recovery from undersampled measurements. Learning such a large network reliably requires extensive amounts of training time and training data, which is often scarce in the biomedical imaging setting. In addition, since the learned inverse network is tied to the specific forward model, different large trained models are often needed in a clinical setting, where different acquisition parameters (e.g. matrix size, resolution, undersampling patterns) may be used.
  The main difference of the proposed scheme with \cite{zhang2017magazine,Zhang2017plugplaycnn,oneNetwork2017iccv} is the use of end-to-end training, similar to \cite{schlemper2017cascadeRueckert, hammernik}. Since the network parameters in the end-to-end training strategy are trained for the specific task of image recovery, it provides a significant improvement in performance over the use of pre-trained denoisers. Specifically, the CNN parameters are learned to capture the alias artifacts and noise at each of the iterations, which depend on the forward model; the customization of the network to the specific task provides improved recovery compared to a generic CNN denoiser. In addition, the proposed scheme does not need a recipe for choosing the noise variance at each iteration or the regularization parameter as in
 \cite{zhang2017magazine,Zhang2017plugplaycnn}. The regularization parameter is learned, while the network is trained to remove the alias patterns and noise at all iterations, which may differ in its statistical properties. A key difference of our approach with \cite{schlemper2017cascadeRueckert,
  hammernik} is that we use the same CNN blocks in all the iterations. Since different CNN blocks are used at each iteration in \cite{schlemper2017cascadeRueckert, hammernik}, these schemes are not
strictly equivalent to the model-based framework. In addition, the proposed scheme offers improved performance over the above schemes in a training data constrained setting. Specifically, several iterations are often needed for the convergence of the variational criterion. When the weights are not shared in \cite{schlemper2017cascadeRueckert, hammernik}, the number of free parameters in these frameworks grows linearly with iterations. By contrast, the number of network parameters do not grow with the number of iterations in our setting. The decoupling of the convergence and the complexity of the network allows us to choose the number of parameters depending on the available training data.  Our experiments show that this approach translates to improved reconstruction performance, especially when training data is limited. 
The main difference of the proposed framework against  proximal gradients methods, which alternate between CNN blocks and a steepest descent step\cite{Mardani2018cvpr,hammernik,putzky,lista}, is the use of conjugate gradients optimization algorithms within the network. Specifically, the proximal gradient algorithm is attractive in compressed sensing (CS) setting since each of the sub-steps are computationally inexpensive, even though the total number of iterations may be high. Unfortunately, the number of iterations/unfolding steps that the proximal gradient RNN can be trained with is limited, especially on GPUs with limited onboard memory. By contrast,  the CG sub-blocks in our approach result in quite a significant reduction in data consistency error at each iteration, thus offering a faster reduction of cost per iteration. Note that replacing a steepest descent step with several CG steps is not associated with increased memory demand during training. Our experiments show that this strategy offers improved performance compared to the proximal gradients based approaches. 

In summary, the main novelties of the proposed scheme over related deep-learning schemes described above are \textbf{(a)} the architecture involving numerical optimization blocks within the deep network, which enables the easy use of complex forward models and additional image priors\cite{modlStormICASSP2018}, while offering better performance than current RNN methods that rely on proximal gradients \cite{Mardani2018cvpr,hammernik,putzky}. The use of numerical optimization blocks within deep networks has not been reported before, to the best of our knowledge. \textbf{(b)} the variational model based formulation, and the iterative algorithm, which translates to an unrolled network architecture with shared weights. By decoupling the network complexity from convergence, the weight sharing strategy provides better performance in training data constrained settings compared to \cite{schlemper2017cascadeRueckert, hammernik}, while the end-to-end training offers improved performance compared to pre-trained models \cite{zhang2017magazine,Zhang2017plugplaycnn,oneNetwork2017iccv}. While some of these components have been independently used  by other researchers, the specific combination naturally emerges from the model-based formulation; in addition, the benefit of this specific strategy over competing methods have not been rigorously tested and validated, which is a contribution of this work.

\section{Background}
\subsection{Image formation \& forward model}
The imaging system can be thought of as an operator $\mathcal A$ that acts on a continuous domain image $x: \mathbb R^2 \rightarrow \mathbb C$ to yield a vector of measurements $\mathcal A(x)=\mathbf b \in \mathbb C^N$. The goal of image reconstruction is to recover a discrete approximation, denoted by the vector $\mathbf x\in \mathbb R^{p}$ from $\mathbf b$. 

Model-based imaging schemes~\cite{ongie2015super,gregtsp2017} use a discrete approximation of $\mathcal A$, denoted by the matrix $\mathbf A$, that maps $\mathbf x$ to $\mathbf b$; model-based algorithms make the central assumption that
\begin{equation}
\mathbf b = \mathcal A(\mathbf x).
\end{equation}
For example, in the single-channel Cartesian MRI acquisition setting,
we have $\mathcal A =  \mathbf S\mathbf F$, where $\mathbf F$ is the 2-D discrete Fourier transform, while $\mathbf S$ is the fat sampling matrix that pick rows of the Fourier matrix. 

The recovery of $\mathbf x$ from $\mathbf b$ is ill-posed, especially when $\mathbf A$ is a rectangular matrix. The general practice in model-based imaging is to pose the recovery as a regularized optimization scheme: 
\begin{equation}
  \mathbf x = \arg \min_{\mathbf
   x}\underbrace{\|\mathbf A\,\mathbf x-\mathbf b\|_2^2}_{\mbox{data
     consistency}} + {\lambda}~\underbrace{\mathcal R(\mathbf x)}_{\mbox{regularization}}
\end{equation}
The regularization prior $\mathcal R: \mathbb C^n \rightarrow \mathbb R_{> 0}$ is often carefully engineered to restrict the solutions to the space of desirable images. For example, $\mathcal R(\mathbf x)$ is a small scalar when $\mathbf x$ is a noise and artifact-free image, while its value is high for noisy images. Classical choices include norms of wavelet coefficients~\cite{ista2003wavelet}, total variation~\cite{hdtv2012jacob}, as well as their combinations. Recently, several authors have also  recently introduced structured low-rank based priors that encourage super-resolution image recovery \cite{gregtsp2017,gregSIAM2016,alohaLee2016,Haldar2014loraks}. Plug-and-play approaches that also rely on off-the-shelf image denoisers have been introduced as regularizers \cite{bm3d}.

\subsection{Deep learned image reconstruction: the state-of-the-art}

Many of the current deep learning based algorithms recover the images as
\begin{equation}
\mathbf x_{\rm rec} = \mathcal T_{\mathbf w} \left(\mathbf A^H \;\mathbf b\right),
\end{equation}
where $\mathcal T_{\mathbf w}$ is a learned CNN~\cite{barauniukDeepInverse2017}. The operator
$\mathbf A^H (\cdot)$ transforms the measurement data to the image domain, since CNNs are designed to work in the image domain. We thus have the relation 
\begin{equation}
\mathbf x_{\rm rec} = \mathcal T_{\mathbf w} \left(\mathbf A^H \mathbf A  \mathbf x\right),
\end{equation}
Thus, the CNN network is learned to invert the normal operator $\mathbf A^H \mathbf A$; i.e., $\mathcal T_{\mathbf w} \approx \left(\mathbf A^H \mathbf A\right)^{-1}$ for signals living in the image set. 

For many measurement operators (e.g Fourier sampling, blurring,
projection imaging), $\mathbf A^H \mathbf A$ is a
translation-invariant operator; the convolutional structure makes it
possible for CNNs to solve such problems \cite{unser2017}. However,
the receptive field of the CNN has to be comparable to the support of the point spread function corresponding to $\left(\mathbf A^H \mathbf A\right)$. In applications involving Fourier sampling or projection imaging, the receptive field of the CNNs has to be the same as that of the image; large networks such as UNET with several layers are required to obtain such a large receptive field. A challenge with such large network with many free parameters is the need for extensive training data to reliably train the parameters. Another challenge is that the CNN structure may not be well-suited for problems such as parallel MRI, where $\mathbf A^H \mathbf A$ is not translational-invariant. 

An alternate approach is to unroll an iterative algorithm involving a
CNN-based regularizer
\cite{Schlemper2017ruckertDynamic,Diamond2017,Zhang2017plugplaycnn}, which  is similar to the proposed scheme; we will discuss the differences between these schemes and the proposed method in the next section.

\section{Proposed Method}
We formulate the reconstruction of the image $\mathbf x \in \mathbf C^n$ as the optimization problem:
 \begin{equation}
 \label{maineqn} \mathbf x_{\rm{rec}} = \arg \min_{\mathbf
   x}\underbrace{\|\mathcal A(\mathbf x)-\mathbf b\|_2^2}_{\mbox{data
     consistency}} + {\lambda}~\underbrace{\|\mathcal N_{\mathbf
     w}(\mathbf x)\|^2}_{\mbox{regularization}}.
\end{equation}
Here, $\mathcal N_{\rm w}$ is a learned CNN estimator of noise and alias patterns, which depends on the learned parameters $\mathbf w$. We express $\mathcal N_{\mathbf w}(\mathbf x)$ as
\begin{equation}
\label{denoising}
  \mathcal N_{\mathbf w} (\mathbf x) = \left(\mathcal I -\mathcal D_{\rm w}\right)(\mathbf x) = \mathbf x-\mathcal D_{\mathbf w}(\mathbf x).
\end{equation}
where $ \mathcal D_{\mathbf w} (\mathbf x)$ is the \emph{"denoised"} version of $\mathbf x$, after the removal of alias artifacts and noise. 
The use of the CNN-based prior $\|\mathcal N_{\mathbf w}(\mathbf x)\|^2$, which gives high values when $\mathbf x$ is contaminated with noise and alias patterns, results in solutions that are data-consistent and are minimally contaminated by noise and alias patterns. Here, $\lambda$ is a trainable regularization parameter. 
Substituting from \eqref{denoising}, in \eqref{maineqn}, we obtain
\begin{equation}
 \label{maineqn2} \mathbf x_{\rm{rec}} = \arg \min_{\mathbf x}\|\mathcal A(\mathbf x)-\mathbf b\|_2^2~ + {\lambda}~\|\mathbf x - \mathcal D_{\mathbf w}(\mathbf x)\|^2
\end{equation}

Since these schemes rely on forward models, the receptive field of the networks need not be the full image size. In addition, since the network only needs to capture the redundancies in the images, a network consisting of many fewer parameters is sufficient to obtain good results. Note that the above formulation is very similar to the plug-and-play prior approach in \cite{chan2017plug}; the main difference is the denoiser is a deep CNN in our setting, similar to~\cite{Zhang2017plugplaycnn}. Unlike \cite{Zhang2017plugplaycnn}, that uses networks pre-trained for denoising, we rely on end-to-end training as described in the next subsection. We set $\lambda$ as a trainable parameter. If the constrained setting can yield improved reconstructions, high values of $\lambda$ would be selected during the training process. 
\begin{figure}
  \centering
 \subfloat[The Residual learning based denoiser]{     
    \centering
    \includegraphics[width=.6\linewidth] {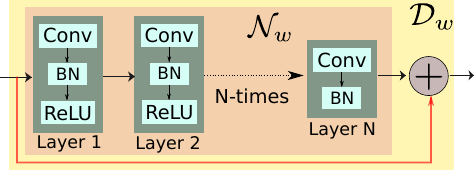}
    \label{subfig:cnn}
  }

  \vspace{.3cm}
\subfloat[Proposed Model-based Deep Learning~(MoDL) architecture]{
    \includegraphics[width=.9\linewidth] {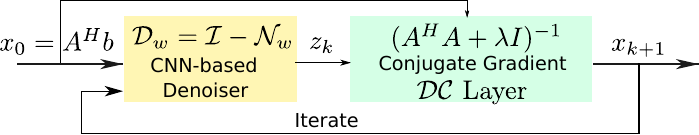}
        \label{subfig:mainarch}
    }
  
  \vspace{.3cm}
\subfloat[Unrolled architecture as described in Eq.~4 and 5.]{
  \includegraphics[width=\linewidth]{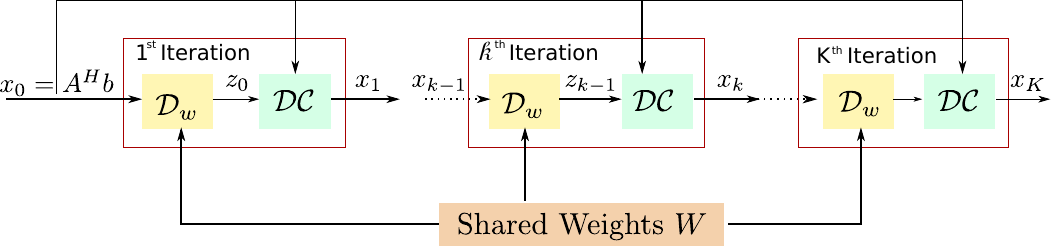}
      \label{subfig:unrolled}
    }
  
  \caption{MoDL: Proposed MOdel-based Deep Learning framework for
    image reconstruction. (a) shows the CNN based denoising block
    $D_w$. (b) is the recursive MoDL framework that alternates between
    denoiser $\mathcal D_w$ in \eqref{eq:alternateStrateg} and the
    data-consistency~(DC) layer in \eqref{dc}. (c) is the unrolled
    architecture for $K$ iterations. The denoising blocks $\mathcal D_w$ share    the weights across all the $K$ iterations.  }
  \label{fig:modl}
\end{figure}

\subsection{Unrolling the recursive network}
We note that the non-linear mapping $\mathcal D_{\mathbf w}\left(\mathbf x_n+\Delta \mathbf x\right)$ can be approximated using Taylor series around the $n^{\rm th}$ iterate as 
\begin{equation}
\mathcal D_{\mathbf w}\left(\mathbf x_n+\Delta \mathbf x\right) \approx \underbrace{\mathcal D_{\mathbf w}\left(\mathbf x_n\right)}_{\mathbf z_n} + \mathbf J_{n}^T\nabla \mathbf x,
\end{equation}
where $\mathbf J_n$ is the Jacobian matrix. Setting $\mathbf x_n+\Delta \mathbf x=\mathbf x$, the penalty term can be approximated as
\begin{equation}
\|\mathbf x - \mathcal D_{\mathbf w}(\mathbf x_n+\nabla \mathbf x)\|^2 \approx \|\mathbf x-\mathbf z_n\|^2 + \| \mathbf J_{n}\Delta \mathbf x\|^2
\end{equation}
We note that the second term tends to zero for small perturbations (i.e, a small value of $\|\nabla \mathbf x\|$). Since the above approximation is only valid in the vicinity of $\mathbf x_n$, we obtain the alternating algorithm that approximate~\eqref{maineqn2}:
\begin{subequations}
  \begin{align}
    \mathbf x_{n+1} &= \arg \min_{\mathbf x}\|\mathcal A(\mathbf x)-\mathbf b\|_2^2~ + {\lambda}~\|\mathbf x - \mathbf z_n\|^2,  \label{maineqn3}\\ 
    \mathbf z_{n} &= \mathcal D_{\mathbf w}\left(\mathbf x_{n}\right)
    \label{eq:alternateStrateg}
\end{align}
\end{subequations}
The sub-problem \eqref{maineqn3} can be solved using the normal equations : 
\begin{equation}
\label{dc}
 \mathbf x_{n+1} = \underbrace{\left(\mathcal A^H\mathcal A + \lambda \;\mathcal I\right)^{-1}}_{\mathcal Q} \left(\mathcal A^H (\mathbf b) + \lambda~ \mathbf z_{n}\right)
\end{equation}
The algorithm is initialized with $\mathbf z_0 = \mathbf 0$. The outline of the iterative framework is shown in Fig.~\subref*{subfig:mainarch}. Once the number of iterations is fixed, the update rules can be viewed as an unrolled deep linear CNN, as shown in Fig.~\ref{subfig:unrolled}, whose weights at different iterations are shared. The proposed unrolled architecture has similarities to \cite{schlemper2017cascadeRueckert,hammernik,Zhang2017plugplaycnn}. However, unlike these works, we use the same denoising  operator $\mathcal D_{\mathbf w}$ at each iteration. Similarly, we use the same trainable regularization parameter at each iteration for consistency with \eqref{maineqn3} \& \eqref{eq:alternateStrateg}.    \cite{schlemper2017cascadeRueckert} and \cite{hammernik},~\eqref{maineqn}.  The key benefit of the proposed scheme is the quite significant reduction in the number of network parameters. Specifically, the number of parameters is smaller by a factor of the number of iterations. Our experiments demonstrate the improvement in the robustness of the training procedure and the improved quality of the reconstructions.  

We relied on using a penalized formulation and an alternating minimization algorithm for simplicity. An alternative is a constrained setting, where the prior is imposed as a constraint $\mathbf x = \mathcal D_{\mathbf w}(\mathbf x)$ using an ADMM scheme \cite{chan2017plug}. In this setting, rigorous convergence guarantees can be derived as in \cite{chan2017plug}. 
Similar rigorous ADMM based architectures are also considered in \cite{admmnet}, which can result in a slightly different architecture; these architectures can be adapted to our setting as well. We note that the weights at different layers are not shared in \cite{admmnet}. We note that convergence guarantees are not too relevant in our setting since it is not practical to iterate the network till convergence.  Note that the constraint in our case is satisfied when $\lambda\rightarrow \infty$. 

 \subsection{Data consistency layer}
 When the forward model is simple, the solution to \eqref{dc} can be analytically computed. For example, $\mathcal A = \mathbf S \mathbf F$ in the single channel undersampled MRI acquisition, where $\mathbf S$ is a sampling matrix, obtained by keeping only the relevant rows of an identity matrix. Specifically, the $m^{\rm th}$ row is retained is the corresponding Fourier sample is sampled. $\mathbf F$ is the Fourier matrix. In this case, the Fourier transform of the image at the sampling location $m$ can be analytically evaluated as 
 \begin{equation}
\widehat {\mathbf x_k}[m] = \left\{\begin{array}{ccc}
\frac{\left(\mathbf b[m]+\lambda \widehat {\mathbf z_k}[m]\right)}{1+\lambda} & \mbox{if } & \mbox{$m^{\rm th}$ sample is acquired}\\\\
\widehat {\mathbf z_k}[m] & \mbox{else. } 
\end{array}
\right.
\end{equation}

The operator $\left(\mathcal A^H\mathcal A + \lambda \;\mathcal I\right)$ is not analytically invertible for complex operators such as multichannel MRI. In this case, we propose to solve \eqref{dc} using conjugate gradient optimization scheme. This implies that the unrolled deep network will have sub-blocks consisting of numerical optimization layers. The standard approach to dealing with complex forward models is to use  proximal gradient algorithms \cite{Mardani2018cvpr,hammernik,putzky}, which alternates between CNN blocks and steepest descent steps. We observe that such algorithms are attractive in compressed sensing (CS) setting since each of the sub-steps are computationally inexpensive, even though the total number of iterations may be high. Unfortunately, the number of iterations/unfolding steps that the proximal gradient RNN can be trained with is limited, especially on GPUs with  limited on-board memory. Specifically, the RNN need to be unrolled for training, a assuming fixed number of iterations; all of the intermediate results at each iteration and layer need to stored to perform backpropagation, which constrains the number of iterations the RNN can be unrolled during training. By contrast, the CG sub-blocks, involving several CG steps, in our approach results in the more accurate enforcement of the data-consistency constraint at each iteration, thus offering a faster reduction of cost per iteration. Note that there are no trainable parameters within the CG algorithm, and hence the intermediate results at each step of the CG algorithm need not be stored to perform backpropagation. The gradients can be backpropagated through the CG block using another CG as shown in Section \ref{endtoend}. This implies that a large number of CG steps can be performed at each iteration with almost no memory overhead during training.  Our experiments show that this strategy offers improved performance compared to the proximal gradients based approaches. In addition to accounting for complex forward models, this approach also facilitate the easy incorporation of other image regularization terms (e.g subject specific priors\cite{modlStormICASSP2018}). 
 
 \subsection{End-to-end training of the deep network}
\label{endtoend}
One strategy for incorporating the deep learned prior in Eq.~\eqref{eq:alternateStrateg} is
to reuse networks pre-trained for image  denoising task in the
reconstruction framework, with a heuristic approach to select the
noise level at each iteration~\cite{Zhang2017plugplaycnn}. Decoupling the training process from the specifics of the acquisition will significantly simplify the approach. However, the statistics of the artifacts introduced by undersampling cannot be fully captured by Gaussian noise process. Our experiments show the performance of the network can be improved significantly by training the network in an end-to-end fashion. 

We fix the number of iterations as $K$ and specify the loss function as the mean square error between $\mathbf x_K$ and the desired image $t$:
\begin{equation}
\mathcal C = \sum_{i=1}^{\rm Nsamples} \|\mathbf x_K(i) - \mathbf t(i)\|^2,
\end{equation}
where $t(i)$ is the $i^{\rm th}$ target image. 

The goal of training is to determine the weight parameter $\mathbf w$, which is shared across the iterations. 
The gradient of the cost function with respect to the shared weights can be determined using the chain rule
\begin{equation}
(\nabla_{\mathbf w} \mathcal C) = \sum_{k=0}^{K-1} \mathbf J_{\mathbf w}(\mathbf z_k)^T \;(\nabla_{\mathbf z_k} \mathcal C),
\end{equation}
where the Jacobian matrix $\mathbf J_{\mathbf w}(\mathbf z)$ has entries $[\mathbf J_{\mathbf w}(\mathbf z)]_{i,j} = \partial z_i/\partial w_j$ and  $\mathbf z_k$ is the output of the CNN at the $k^{\rm th}$ iteration.

Note that the noise/alias terms in the signal at each of the iterations $\mathbf x_k$ may be significantly different. In plug-and-play CNN approaches, different networks that are trained for (pre-determined) decreasing noise variances \cite{zhang2017magazine,Zhang2017plugplaycnn} are used at different iteration.  By contrast, we share the same network $\mathcal D_{\mathbf w}$ and same regularization parameter $\lambda$ across iterations. During training, the non-linear CNN thus learns to denoise noise/alias patterns with several different statistics. The added benefit of our end-to-end training approach is that we do not require a recipe for the choosing the noise variance at each iteration or regularization parameter.

Note that the backpropagation scheme requires the evaluation of the terms $\nabla_{\mathbf z_k} \mathcal C, k=0,..,K-1$. These terms can be evaluated recursively using backpropagation as in a regular deep learning network. The main difference with the traditional CNN training is that we have numerical optimization blocks within the deep network. We now focus on how to back-propagate through these conjugate gradient (CG) blocks \cite{numerical_recipes_c}. Note that the CG block does not have any trainable parameters. We have 
\begin{equation}
\nabla_{\mathbf z_{k-1}} \mathcal C = \mathbf J_{\mathbf z_{k-1}}(\mathbf x_k)^T~ \nabla_{\mathbf x_k} {\mathcal C}
\end{equation}
where the Jacobian matrix $\mathbf J_{\mathbf z}(\mathbf x)$ has entries $[\mathbf J_{\mathbf z}(\mathbf x)]_{i,j} = \partial x_i/\partial z_j$. Note from \eqref{dc} that $\mathbf x_{k} = \mathcal Q \,\mathbf z_{k-1} + \mathbf q $, where $\mathcal Q=\left(\mathcal A^H\mathcal A + \lambda \;\mathcal I\right)$ and $\mathbf q = \mathcal A^H (\mathbf b)$. The Jacobian matrix of $\mathcal Q$ is given by
\begin{equation}
\mathbf J_{\mathbf z}(\mathbf x) = \left(\mathcal A^H\mathcal A + \lambda \;\mathcal I\right)^{-1}
\end{equation}

Since $\mathcal Q$ is symmetric, we have

\begin{equation}
\label{backprop}
(\nabla_{\mathbf z_{k-1}}\mathcal C)  = \underbrace{\left(\mathcal A^H\mathcal A + \lambda \;\mathcal I\right)^{-1}}_{\mathcal Q}~ (\nabla_{\mathbf x_k} \mathcal C)
\end{equation}
We can evaluate the above expression using a CG algorithm, run until convergence, determined by the saturation of the data-consistency cost. Note that the above gradient calculation is only valid if we implement $\left(\mathcal A^H\mathcal A + \lambda \;\mathcal I\right)^{-1}$ or equivalently let CG algorithm converge. This result thus shows that the gradients can be back-propagated through the CG block using a CG algorithm. Once the relation \eqref{backprop} is established, we propose to update the network parameters $\mathbf w$ of the network $\mathcal D_{\mathbf w}$ as well as the regularization parameter
$\lambda$ using the Adam optimization scheme~\cite{Kingma2015}.  This scheme maintains two learning rates corresponding to each parameter. These learning rates are estimated from the first and second moments of the gradients. Note
that the data consistency term specified by \eqref{dc}, and hence its
gradient can be computed analytically in the Fourier domain. This
strategy allows the computation of the gradients of the weights using
backpropagation; the weight gradients evaluated using the chain rule
will have contributions from all the iterations. To make the learned
network less sensitive to changes in acquisition scheme, we use training
data with different undersampling patterns. We rely on the variable
sharing strategies in TensorFlow to implement the unrolled
architecture. The parameters of the networks at each iteration are initialized and updated together. 

\subsection{Implementation details}
 
The CNN architecture used in this work is shown in
Fig.~\ref{subfig:cnn}. We used $N$~layer model with 64 filters at each
layer to implement the $\mathcal N_w$ block.  Each layer consists of
convolution~(conv) followed by batch normalization~(BN)~\cite{Ioffe2015BN} and a
non-linear activation function ReLU~(rectified linear unit, 
$f(x)=\max(0,x)$). The $N^{th}$-layer does not have ReLU to
avoid truncating the negative part of the learned noise patterns.
Following the residual learning strategy, the learned noise from
$\mathcal N_w$ block is added with the input of $\mathcal N_w$ block to get the reconstructed
image as the output of the $\mathcal D_w$ block.  The number of
trainable parameters at each layer of the CNN network is shown in
Table \ref{tab:parameters}.   The output of
$\mathcal D_w$ block is fetched into data consistency~(DC) layer as
described in Fig.~\ref{subfig:mainarch}.  The proposed
recursive model, shown in Fig.~\ref{subfig:mainarch}, was  unrolled
assuming $K$~iterations of the alternating
strategy~\eqref{eq:alternateStrateg} and implemented in
TensorFlow. Specifically, we set the number of layers $N$ as $5$ and
number of iterations $K$ as $10$.
Since MR
images are complex, the data consistency~(DC) layer explicitly works with
complex inputs and returns a complex output. The CNN part handles
complex data by concatenating the real and imaginary part as channels
i.e. we convert from $\mathbb{C}^{m\times n}$ space into  $
\mathbb{R}^{m\times n\times2}$ space.  

We use a two-step approach to initialize the network for training. We first trained a model for only one iteration, initializing the parameters with random values. This training is considerably faster than training the entire network. Since we use a recursive network with the same weights across iterations, the weights of the unrolled network at each iteration in Fig.~\ref{subfig:unrolled} are initialized using the weights learned from the single iteration model. We observe that this two-step training strategy is considerably faster and reliable than initializing the full network using random weights. Since we use the same weights across iterations, we may choose the number of iterations in the reconstruction algorithm to be different from the ones assumed during training. The source code for the proposed MoDL scheme can be downloaded from this link: \texttt{\url{https://github.com/hkaggarwal/modl}}   
\begin{table}[t!]
\caption{Description of the trainable parameters used in the  deep network in case of with sharing~(WS) and no sharing~(NS)
  architectures between iterations. The weight sharing strategy provides a 10~times reduction in the number of trainable parameters, improving robustness when training data is scarce.}
\label{tab:parameters}
\begin{tabular}{cccc}  \toprule
          & BN ($\beta+\gamma+\mu+\sigma^2$)& Conv.filters  & Total  \\ \midrule
conv1     & $64+64+64+64$        & $ 3\times 3 \times 2 \times 64  $  & 1408    \\
conv2     & $64+64+64+64$        & $ 3\times 3 \times 64\times 64  $  & 37120   \\
conv3     & $64+64+64+64$        & $ 3\times 3 \times 64\times 64  $  & 37120   \\
conv4     & $64+64+64+64$        & $ 3\times 3 \times 64\times 64  $  & 37120   \\
conv5     & $2+2+2+2    $        & $ 3\times 3 \times 64\times 2   $  & 1160    \\
$\lambda$ &                      &                 & 1        \\  \midrule
  \multicolumn{3}{c}{number of trainable parameters in WS strategy:} &113,929 \\
  \multicolumn{3}{c}{no.~of parameters in NS  $=$ \#WS $\times$ 10 iterations:} & 1,139,290 \\ \bottomrule
\end{tabular}
\end{table}

\section{Experiments and Results}
This section describes various experiments conducted to quantitatively and qualitatively evaluate the performance of the proposed method.
\subsection{Deep learning variants used in the validation}
\begin{table}
 \caption{Categorization of iterative deep-learning frameworks} 
 \label{tab:dlFrameworks}
\renewcommand{\tabularxcolumn}[1]{>{\arraybackslash}m{#1}}
\newcommand{\scell}[2][c]{%
  \begin{tabular}[#1]{@{}c@{}}#2\end{tabular}}
  \centering
  \begin{tabularx}{\linewidth}{@{\hspace{.2cm}}c@{\hspace{.3cm}}cc}  \toprule
    \textbf{Optimization Strategy} & \textbf{Training Strategy} & \textbf{Network Architecture} \\ \midrule
    \scell{Steepest Descent\\ (SD)} & \scell{Pre-trained Denoiser \\ (PD)}& \scell{No Sharing \\(NS)}\\[.2cm]
   \scell{ Conjugate Gradient \\(CG)} & \scell{ End-to-End Training \\(ET)} & \scell{With Sharing\\ (WS)}\\[1em] 
  \end{tabularx}
  \begin{tabularx}{\linewidth}{cX} \toprule
    \multicolumn{2}{c}{Deep learning frameworks derived based on    above three parameters} \\ \midrule
    \textbf{Framework } & \multicolumn{1}{c}{\textbf{Description}} \\ \midrule
     CG-ET-WS &  Proposed MoDL framework, which uses conjugate gradient algorithm to solve the DC subproblem, with end-to-end training~(ET) strategy, and with sharing~(WS) of weights across iterations.\\[.3em]
     CG-ET-NS & The difference from MoDL is that the weights are not shared (NS) across iterations.\\[.3em]
     SD-ET-WS & The difference from MoDL is the change the optimization algorithm in the DC layer to steepest descent~(SD) instead of CG. \\[.3em]
     CG-PD-NS & The difference from MoDL is the use of pre-trained denoisers~(PD) within the iterations. We trained 10 different $D_w$ blocks with descending noise variance and use them in 10~iterations during testing. Since these are different denoisers, therefore, the weights are not shared (NS).\\
    \bottomrule
  \end{tabularx}

\end{table}
The implementations of many of the existing deep
  learning methods are not readily available and often not directly applicable in our setting. We hence propose to compare our scheme against variants that  differ on optimization strategy, training approach,
and network architecture as shown in Table.~\ref{tab:dlFrameworks}. All these variants were trained with the same dataset as for MoDL and with the same number of slices (total 360). The
proximal gradients algorithm uses a single steepest descent (SD) step to encourage data consistency at each iteration. By contrast, we solve the quadratic subproblem using CG at each iteration.  We rely on end-to-end training~(ET) of the network to learn the parameters. An alternate strategy in the literature is to use
pre-trained CNN  based
  denoiser~(PD). In  the PD scheme, we use different denoisers at each step, which are
  pre-trained to different noise levels that monotonically decrease
  with iteration as suggested in
  \cite{zhang2017magazine,Zhang2017plugplaycnn}. In \cite{zhang2017magazine,Zhang2017plugplaycnn}, 25 different models were trained with noise levels between 5 and 25 with a step size of 2. Similarly, we had trained 10 different models between noise levels 0.02 and 0.25 with step size of 0.02 (nearly). The particular values were: 0.02, 0.04, 0.06, 0.08, 0.10, 0.13, 0.15, 0.17, 0.20, 0.25.   The
  regularization parameter in the PD setting is tuned to obtain the
  best performance to be fair. The same value of $\lambda$ was used in all the iterations. Based on these parameters the proposed
MoDL framework can  be categorized  as CG-ET-WS strategy.

In the following subsections, we show the benefit of
CG-ET-WS strategy by varying each of these three parameters.  Specifically, we compare with SD-ET-WS
approach where we utilize SD algorithm in DC step instead of CG. We
also compare with CG-PD-NS approach where a different pre-trained denoiser is
utilized at each iteration of
Eq.~\eqref{eq:alternateStrateg}. Further, we compare with  CG-ET-NS
technique where we do not share the weights of CNN across iterations of
Eq.~\eqref{eq:alternateStrateg}.

Since training data are scarce in the medical imaging setting, we relied on a dataset with images of cats and dogs for the experiments that determine the impact of training data size. Please note that we did not use this data to pre-train the network in the remaining experiments involving MRI data. The CatDog dataset is available for research from the Kaggle website
\url{https://www.kaggle.com/c/dogs-vs-cats/data}. This dataset
consists of 25,000 images of cats and dogs. We extracted a random subset of
3,000 images for training and 100 images for testing. The images were
pre-processed to crop and resize to the same spatial dimensions of $256
\times 232$ as that of MR dataset. 
We simulated the MRI acquisition on the CatDog dataset. The complex DFT of each of the images were sampled on a 4x undersampling pattern using variable-density Cartesian random sampling masks. We did not assume multichannel sampling. The goal of using this dataset is to demonstrate the reduced data demand of the proposed algorithm and not to produce state of the art results.

\subsection{Data for training and testing the algorithm}
The MRI data used for this study were acquired using a 3D~T2~CUBE sequence with Cartesian readouts using a $12$-channel head coil. The matrix dimensions were 256 $\times$ 232 $\times$ 208 with $1$~mm isotropic resolution.  Fully sampled multi-channel brain images of five volunteers were collected out of which data from four subjects were used for training, while the data from the fifth subject were used for testing. Since the readout is fully sampled, we evaluated the inverse Fourier transform of each readout.  We retrospectively undersampled the phase encodes to train and test the framework; we note that this approach is completely consistent with a future prospective acquisition, where a subset of phase encodes can be pre-selected and acquired. All the experiments were performed with variable-density Cartesian random  sampling mask with different undersampling factors mentioned at their use.

Out of the total $208$ slices for each
subject,  we selected $90$ slices that had images of parts of the anatomy for
training.   The coil sensitivity maps were estimated from the central
k-space regions of each slice using
ESPIRiT~\cite{espirit2014} and were assumed to be known during experiments. Thus,
the training data had dimensions in $\text{rows} \times \text{columns}
\times \text{slices} \times \text{coils}$ as $256\times 232 \times 360
\times 12$ and testing data had dimensions $256 \times 232 \times 164
\times 12$. The testing was performed on $164$ slices out of $208$
available for the test subject since the initial $22$ and the last $22$ slices did not have any brain region but
noise.
To reduce the sensitivity to acquisition settings,
including undersampling patterns,  we used different variable density Cartesian pseudo-random sampling masks for each training slice.  The same
sampling mask was used for the data from all $12$ coils of the same slice. The undersampling factors are mentioned at their use in the experiments. The sampling masks used during the testing were different from the ones used during the training.

\subsection{Impact of optimization strategy and number of iterations}
\begin{figure}
   \centering
   \begin{tikzpicture}
 \begin{axis}[
   width=.9\linewidth,
   height=2in,
   xtick={1,2,...,10},
   ymin=29,
   ymax=38,
   xlabel=Number of iterations during training,
   ylabel=PSNR (dB) on test data,
   grid=both,   
   grid style={line width=.1pt, draw=gray!10},
   major grid style={line width=.2pt,draw=gray!50},
   minor tick num=1,
   legend style={at={(.97,0.45)},draw=none},
   ]
   \addplot[mark=*,black] table [x=Iteration, y=GD] {gdCG.csv};
      \addlegendentry[text width=1.7cm]{SD-ET-WS}
   \addplot[mark=*,red] table [x=Iteration, y=CG] {gdCG.csv};
      \addlegendentry[text width=1.7cm]{CG-ET-WS (MoDL)}
   
  \end{axis}
\end{tikzpicture}

  \caption{Improvement in reconstruction quality at 10x acceleration
as we increase the number of iterations of the network during
training. We observe that the testing performance  saturates around 8-10
iterations.  }
  \label{plt:trainingIterations}
\end{figure}
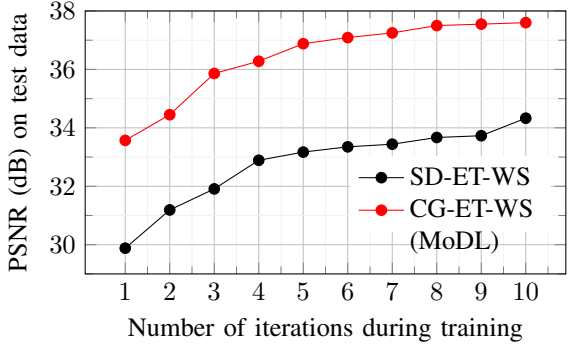

The graph in Fig.~\ref{plt:trainingIterations} shows the effect of increasing the
number of iterations of alternating strategy in
Eq.~\eqref{eq:alternateStrateg}. We compare the proximal gradients approach (SD) with the proposed approach of using CG blocks within the network. The graph also indicates the benefit of utilizing CG instead of SD in the DC block. 
Since CG solves the quadratic subproblem in \eqref{dc} completely as compared to SD, we get   a faster reduction of data consistency cost per CG block; this translates to better performance. Note that there are no trainable parameters within the CG block, and hence no intermediate results need to be stored within this block; the memory demand of the training scheme is only proportional to the number of iterations and is independent of the number of CG steps per iteration. The faster reduction of data-consistency cost translates to improved performance than proximal gradients based schemes with the same number of iterations.

It can be  observed that average PSNR values on the testing data improve as we increase the number of model iterations. Therefore, it is beneficial to unroll the model for several iterations. Since the performance of MoDL saturates around 8-10 iterations, we used this setting in the rest of the experiments. 
We trained 10 different five-layer models corresponding
to each of the 10 iterations as indicated on the x-axis in Fig.~\ref{plt:trainingIterations}. The graph was plotted for average PSNR
values obtained corresponding to the reconstruction at 10x acceleration on
the 164 testing slices.

\subsection{Effect of dataset size}
\begin{figure}
  \centering
  \begin{tikzpicture}
 \begin{axis}[
   width=.95\linewidth,
   height=2.5in,
   xmode=log,
   log ticks with fixed point,
   xtick={.05e3,.1e3,.2e3,.3e3,.6e3, .9e3,2.1e3},
   ytick={22,23,24,25,26,27,28,29,30},
   ymin=22.5,
   ymax=31,
   xlabel=Number of training samples,
   ylabel=PSNR (dB) on testing data,
   grid=both,   
   grid style={line width=.1pt, draw=gray!10},
   major grid style={line width=.2pt,draw=gray!50},
   minor tick num=1,
   legend style={draw=none,
             at={(1,.4)},
              cells={align=left}
            },
   ]
   \addplot[mark=*,blue] table [x=data, y=wsw3L] {catdog1.csv};
    \addlegendentry[text width=2.2cm]{ 3L CG-ET-NS}

    \addplot[mark=*,black] table [x=data, y=wsw5L] {catdog1.csv};
         \addlegendentry[text width=2.2cm]{5L CG-ET-NS }    

    \addplot[mark=*,red] table [x=data, y=sw7L] {catdog1.csv};
    \addlegendentry[text width=2.2cm]{7L CG-ET-WS (MoDL)}

  \node[red,text width=2 cm, font=\footnotesize,align=center] at (axis cs: 100,30.2){ \# parameters: $188$K};
  \node[blue,text width=2 cm, font=\footnotesize, align=center] at (axis cs: 250,28){\# parameters: $199$K};
  \node[black,font=\footnotesize, text width=2cm,align=center] at (axis cs: 1600,27.8){\# parameters: \centering $571$K};
    
  \draw[->,red] (axis cs:130, 30) to[bend right=-50] (axis cs:160, 29.25);
  \draw[->,blue] (axis cs:400, 28) to[bend right=-50] (axis cs:500, 27.7);
  \draw[->,black] (axis cs:1100, 27.5) to[bend right=-30] (axis cs:800, 28.4);
  
  \end{axis}
\end{tikzpicture}

\caption{Effect of training dataset size. The x-axis is shown on the
logarithmic scale. nL in the legend represents that n layer model was used. Figure shows the change  in PSNR  values on the
testing data as we increase the training  dataset size from 50 samples to 2100 samples. }
\label{fig:catdog}
\end{figure}
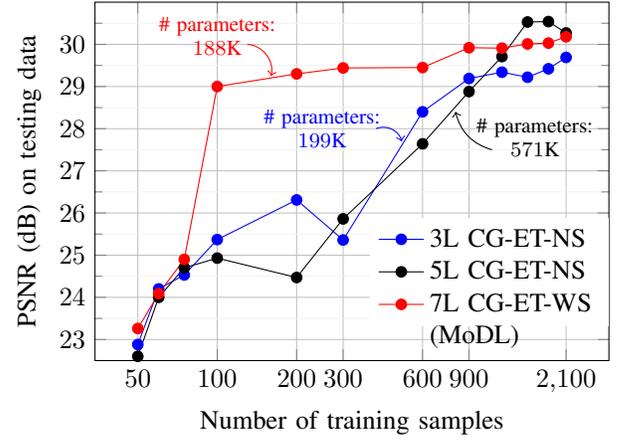

 We relied on the CatDog dataset to determine the impact of training data size on performance. 
Figure~\ref{fig:catdog} shows the comparison of the with sharing~(WS) network
architecture with that of no sharing~(NS) network architecture when the number
 of training images were increased from $50$ to $2100$.  To conduct a fair
comparison of the two network architectures WS and NS, we trained two five-iteration models with nearly the same number of trainable parameters corresponding to WS and NS architectures. In
particular,  a 3~layer~(3L) model  with NS strategy had 199K number of trainable
parameters that is almost same as the number of parameters in the
7~layer WS   strategy model which have 188K parameters.

We observe that our WS model is relatively insensitive to training data beyond $100$ images, thanks to significantly reduced model parameters over NS strategy. Specifically, increasing the number of training samples from $100$ to $2100$ (21 fold increase) only resulted in $1$~dB improvement in PSNR during testing. By contrast, the black curve in Fig.~\ref{fig:catdog} correspond to an NS model with three times more network parameters, which requires significantly more  data to achieve the same performance as the WS scheme. The blue curve corresponds to an NS strategy with the same number of parameters as the WS scheme with five unfolding iterations. The performance of this scheme is worse than the proposed scheme at all training data sizes. This comparison suggests that it is better to use a more complex CNN  and share its weights across iterations when more data is available.

\subsection{Insensitivity of the framework to acquisition settings}
\begin{figure} \centering
  \input{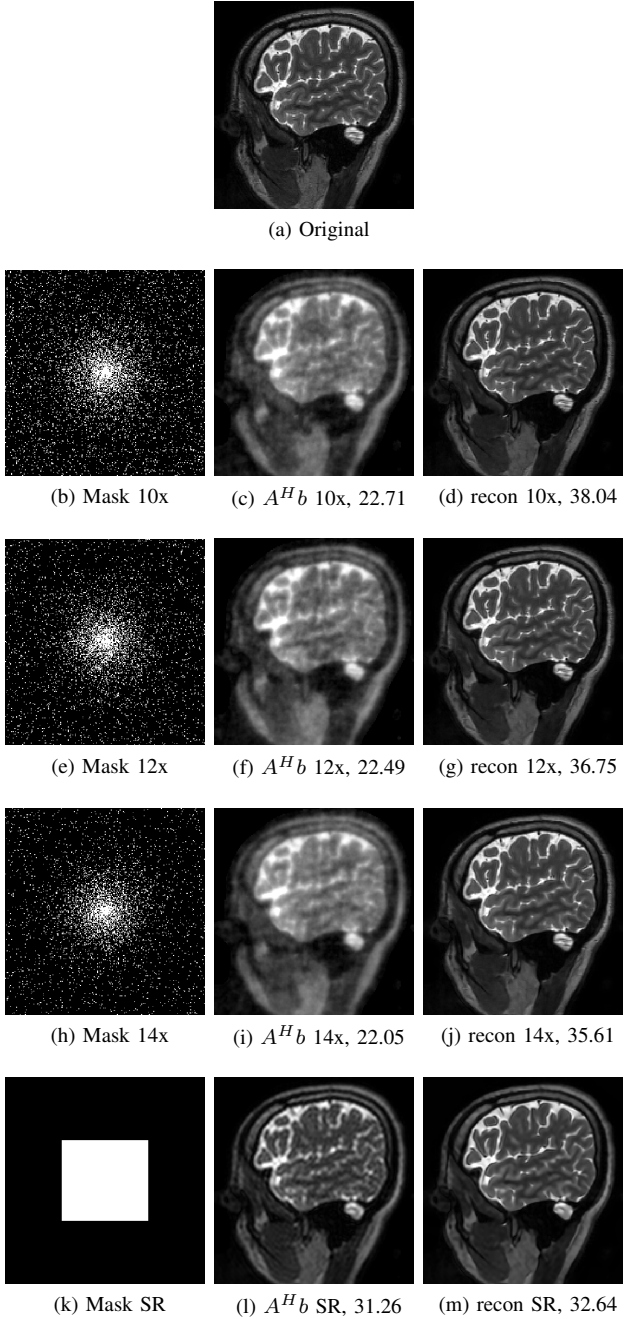}
  \caption{The insensitivity of the trained network to acquisition
    settings.  A single MoDL in a 10-fold   acceleration
    setting was used to recover images from 10x, 12x, and 14x acceleration
    as well as in super-resolution~(SR) settings. The numbers shows PSNR values in dB.}
  \label{fig:higherAcceleration}
\end{figure}

Note that we trained the network with different undersampling patterns to reduce the insensitivity to a specific sampling pattern. In this set of experiments,  we show that a single model trained in a
particular acceleration setting can be utilized to reconstruct the
images from different acceleration settings. In particular, we utilized the trained network at 10x acceleration to reconstruct images
from 10x, 12x, and 14x acceleration settings.    These results in
Figure~\ref{fig:higherAcceleration} show that the trained    network is relatively insensitive to the undersampling setting.
Since we account for the forward model in the network, we expect the
learned network to be relatively insensitive to acquisition
settings. To improve the insensitivity, we had trained the network
with different undersampling patterns with each slice but with same 10x acceleration factor. This training procedure forces the network to learn the manifold of image patches rather than the properties of the forward model; the network learns to estimate the alias patterns and noise that is not localized to the manifold.

Further experiments were carried to check the robustness of the
trained model in super-resolution~(SR) settings. We took the central
$100\times 100$ k-space region as the sampling mask shown in
Fig.~\ref{subfig:srmask}. The resulting low-resolution image
Fig.~\ref{subfig:sratb}  was passed
through the same network, that was trained for 10x acceleration, to get the high-resolution
image as shown in Fig.~\ref{subfig:sr}.

\begin{figure*}
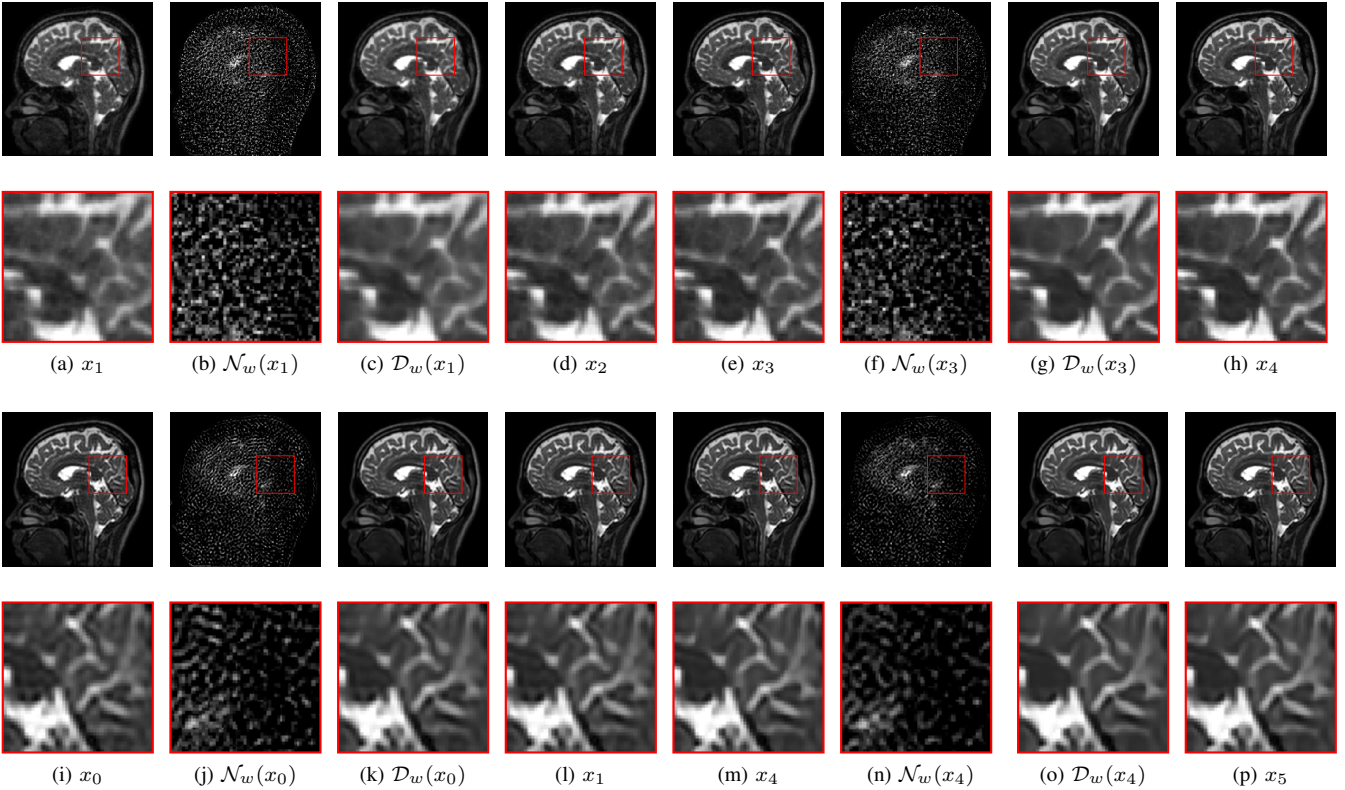

  \input{figIterations16F-2}
  \input{figIterationsSR-2}
\caption{Intermediate results in the deep network. Figure (a)-(h)
  corresponds to the 16-fold acceleration setting, where (a)-(d) corresponds to iteration 2 and (e)-(h) corresponds to iteration 4. Note that the network
  at each iteration estimates the alias and noise signals denoted by
  $\mathcal N_{\mathbf w}(\mathbf x_k)$ from the signal to obtain the
  denoised image $z_k=\mathcal D_{\mathbf w}(\mathbf x_k)$. Figures (i)-(p) corresponds to the super-resolution setting considered in Fig.~\ref{fig:higherAcceleration}. (i)-(l) corresponds to iteration~1 and (m)-(p) corresponds to iterations~5. At $i^{th}$ iteration, $x_{i-1}$ is the input and $x_i$ is the output. Note that the nature of the noise in both cases is very different. Nevertheless, the same network trained at 10x setting is capable of effectively removing the undersampling artifacts. }
\label{fig:srIterations}
\end{figure*}

We show the intermediate steps of the reconstruction scheme at two
different iterations in Fig.~\ref{fig:srIterations}. The top row correspond to the 16-fold random undersampling setting. Please refer to
Fig.~\ref{subfig:unrolled} for an illustration of the unrolled network
architecture. At each iteration, the input $x_k$ is fed to the
plug-and-play CNN, which extracts the alias/noise components $\mathcal
N_{\mathbf w}(x_k)$. We note that the noise/alias terms estimated at each
iteration are different, with the variance within the brain regions
decreasing as iterations progress. The addition of the noise
components to the input yields the denoised output $\mathcal
D_{\mathbf w}(x_k)= \mathcal N_{\mathbf w}(x_k)+x_k$. The data
consistency block, denoted by $\mathcal{DC}$ combines the denoised
outputs with the other terms to yield $x_{k+1}$. The process is
repeated for ten iterations that gradually improve the reconstruction
quality and decrease the noise as evident from the figure.

The end-to-end training strategy ensures that the CNN learns the
information that is complementary to the ones obtained from coil
sensitivity  and data consistency information. Fig.~\ref{fig:srIterations} also shows the reconstruction outputs extracted from intermediate layers in a super-resolution setting. It is
evident from Figures~\ref{subfig:srNoise1} and~\ref{subfig:srNoise5}
that network is able to predict the alias patterns and the aliasing
decays as the number of iterations increase. Note that the same network was used in both the experiments, which shows the ability of the same network to remove alias patterns that are dramatically different.

\subsection{Comparison with other deep learning frameworks}

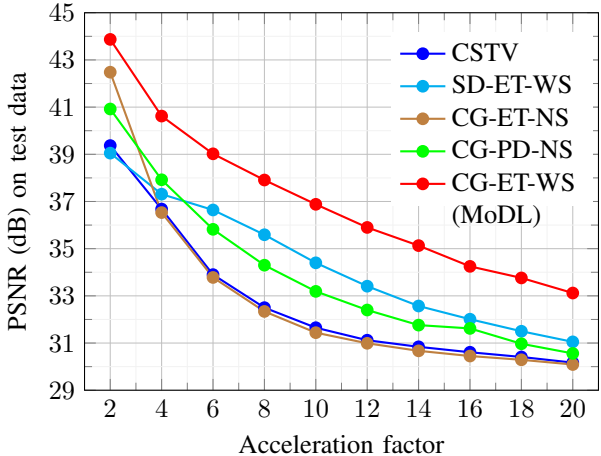
\begin{figure}
  \centering  
  \begin{tikzpicture}
 \begin{axis}[
   width=.95\linewidth,
   height=2.6in,
   xtick={2,4,...,20},
   ytick={29,31,...,45},
   ymin=29,
   ymax=45,
   xmin=1,
   xmax=21,
   xlabel=Acceleration factor,
   ylabel=PSNR (dB) on test data,
   grid=both,   
   grid style={line width=.1pt, draw=gray!10},
   major grid style={line width=.2pt,draw=gray!50},
   minor tick num=1,
   legend columns=1,
   legend style={at={(.99,.95)},draw=none},
   ]
   \addplot[mark=*,blue,thick] table [x=acc, y=cstv,col sep=comma] {comparison.csv};
   \addlegendentry[text width=1.7cm]{CSTV}
   \addplot[mark=*,cyan,thick] table [x=acc, y=Pg-et-ws,col sep=comma] {comparison.csv};
   \addlegendentry[text width=1.7cm]{SD-ET-WS}
   \addplot[mark=*,brown,thick] table [x=acc, y=Cg-et-ns,col sep=comma] {comparison.csv};
   \addlegendentry[text width=1.7cm]{CG-ET-NS}
   \addplot[mark=*,green,thick] table [x=acc, y=Cg-pd-ns,col sep=comma] {comparison.csv};
   \addlegendentry[text width=1.7cm]{CG-PD-NS}
   \addplot[mark=*,red,thick] table [x=acc, y=Cg-et-ws,col sep=comma] {comparison.csv};
   \addlegendentry[text width=1.7cm]{CG-ET-WS (MoDL)}
  \end{axis}
\end{tikzpicture}

  \caption{Performance comparison of various algorithms at different
    acceleration factors. It can be observed that the proposed MoDL
    performs better than other techniques for all different  acceleration factors.}
  \label{plt:comparison}
\end{figure}

In this subsection, we compare the proposed MoDL framework~(CG-ET-WS)
with other three deep learning frameworks as described in
Table~\ref{tab:dlFrameworks}. We also compare the performance of the proposed MoDL
framework with compressed sensing based technique that utilizes total
variation regularization acronym-ed as CSTV~\cite{shiqianma2008}. The
graph shown in Fig.~\ref{plt:comparison} compares the average PSNR
values of 164 slices  on
testing data when acceleration factor varies from 2x to 20x. For deep
learning methods CG-ET-WS, CG-ET-NS, and SD-ET-WS training was
performed on  $10$x acceleration in the presence of Gaussian noise of
standard deviation $\sigma=0.01$ with same amount of training data. The method CG-PD-NS was
trained with uniformly decreasing amount of Gaussian noise from $\sigma=0.25$ to
$\sigma=0.02$ for a total of ten different noise levels. Table~\ref{tab:comparison} shows quantitative values for 6x and 10x acceleration for different algorithms. It can be observed that the standard deviation~(std) of PSNR values from from the mean  is lowest for the proposed MoDL framework. 

\begin{table}
\centering
\caption{Comparison of PSNR~(dB) values obtained by different methods. The values are shown for 6x and 10x acceleration factors in the format: $\mean\;\pm\;std,\; \min/\max$. }
\label{tab:comparison}
\begin{tabularx}{\linewidth}{@{\hspace{.4\tabcolsep}}ccc} \toprule
Method   & 6 fold acceleration      & 10 fold acceleration                 \\ \midrule
CSTV     & $34.50\pm1.62,\;28.42/37.39 $& $31.98\pm1.26,\;29.00/35.26 $\\
SD-ET-WS & $37.30\pm1.42,\;29.44/41.07 $& $34.63\pm1.93,\;30.06/39.51 $\\
CG-ET-NS & $33.83\pm1.97,\;23.33/38.39 $& $31.53\pm1.65,\;24.98/35.84 $\\
CG-PD-NS & $36.52\pm1.72,\;29.57/39.93 $& $33.89\pm1.35,\;30.03/38.03 $\\
MoDL     & $39.24\pm1.18,\;35.27/42.38 $& $37.35\pm1.16,\;32.70/40.61 $\\ \bottomrule
\end{tabularx}
\end{table}

\begin{figure*} \centering
    \input{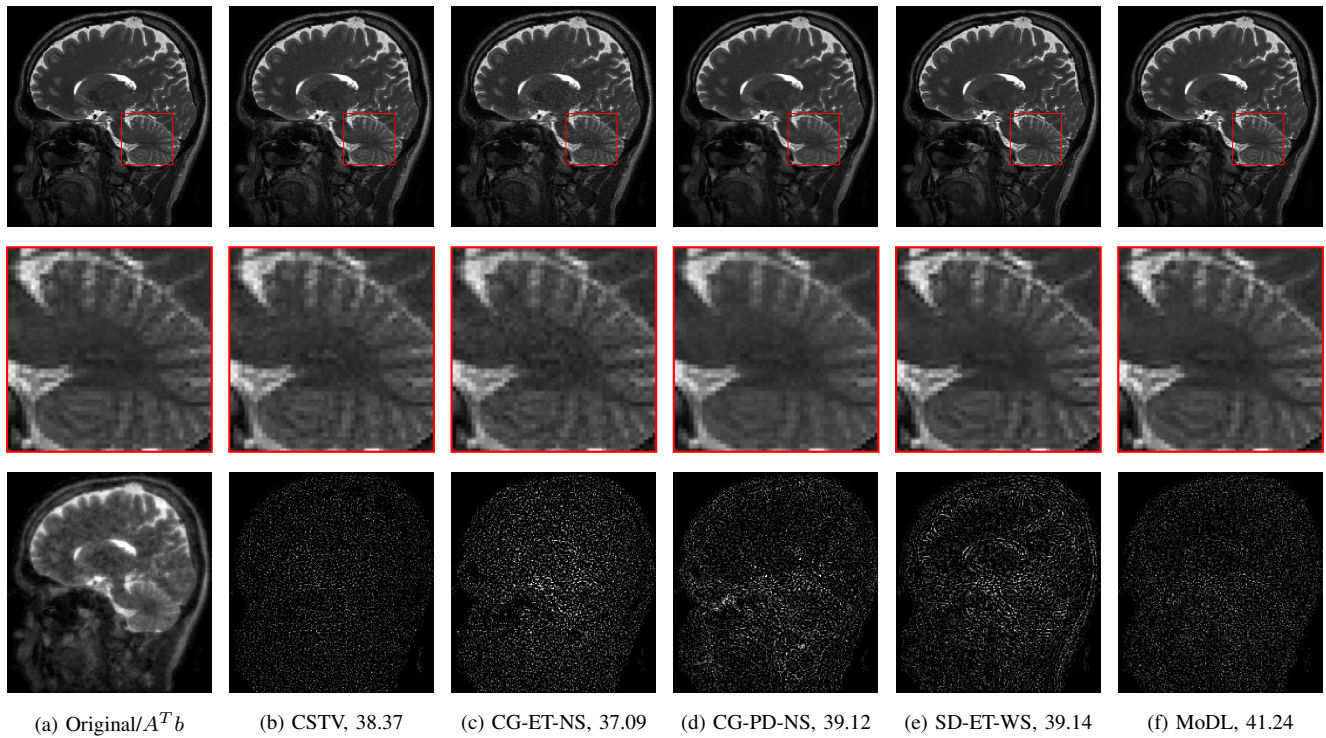}
  \caption{Comparison of the proposed MoDL framework with state of the
    art parallel imaging strategies. The experiments correspond to a 4-fold accelerated case with random noise of $\sigma=0.01$ added
    in k-space. The column 1 shows fully sampled image on top and
    $A^Hb$ on the bottom. The row~2 shows zoomed portions of the
    reconstructed images by different methods. The row~3 shows
    corresponding error images.  The numbers in the caption shows the PSNR values in
    dB. The $A^Hb$ had PSNR value of $25.30$~dB.
    }
  \label{fig:comparison6f}
\end{figure*}

The consistently improved performance of proposed MoDL for different
acceleration factors may be attributed to the combination of the CG in
DC layer, end-to-end training of unrolled network, and a weight
sharing network architecture.  Further, the optimization is performed
on a much lower dimensional space, which is more robust when training
data is scarce.  Since the Gaussian noise model does not capture the
complexity of the alias-induced noise, the PSNR of the CG-PD-NS
network is lower than the proposed one. These results show the benefit
of performing end-to-end training, compared to pre-training strategy.

The technique CG-ET-NS performs poorest among all other methods since
an NS network architecture is a 10 times large capacity model than
corresponding weight sharing strategy. Therefore, it requires a large
amount of training data to learn the aliasing patterns. The proposed
MoDL performs better than compressed sensing based technique CSTV
because the MoDL has the benefit of adaptive learning the regularization-prior from the data itself as opposed to fixed total variation prior used in CSTV. Note that CG-PD-NS provides better performance than CG-ET-NS since the denoisers at each iteration are trained independently to denoise from Gaussian corrupted images, and hence it does not suffer from training data constraint unlike the end-to-end strategy in CG-ET-NS. Significantly more training data is needed to obtain good performance with CG-ET-NS as seen from Fig.~\ref{fig:catdog}. 

\begin{figure*} \centering
  \input{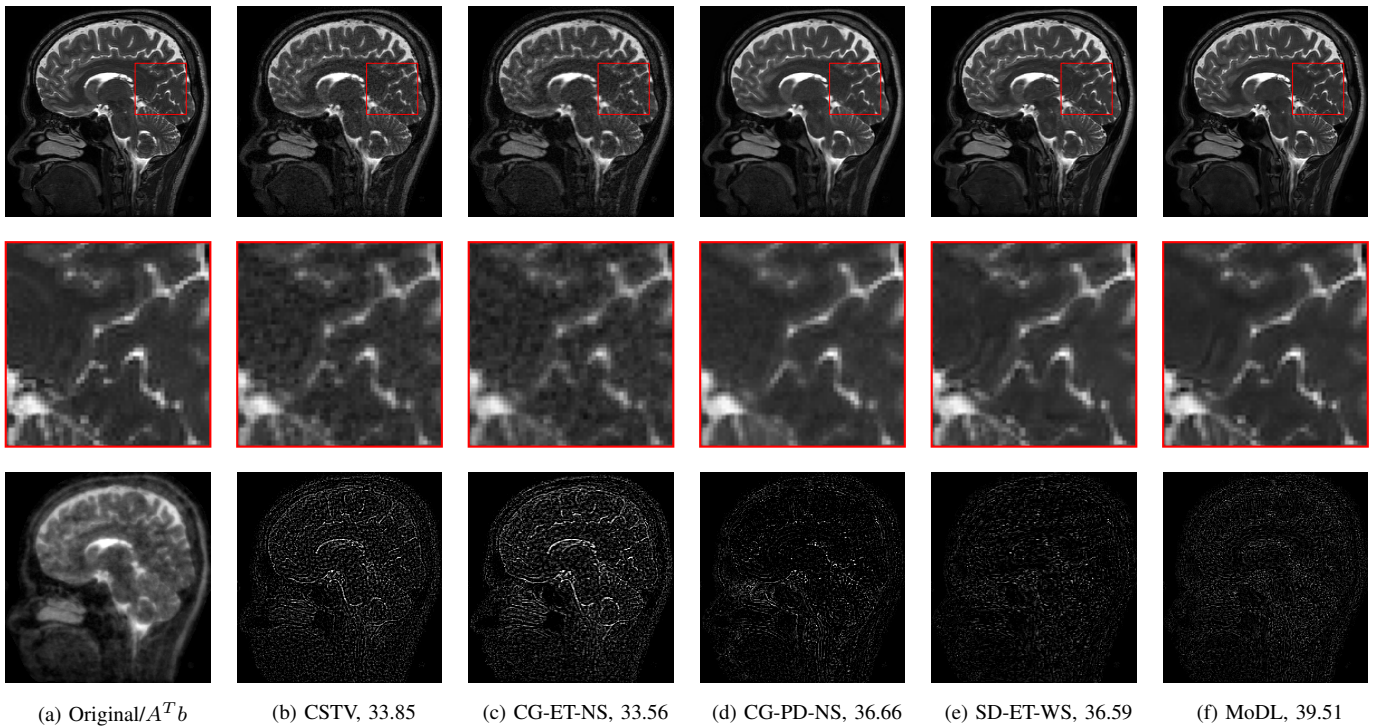}
  \caption{Comparison of the proposed MoDL framework with state of the
    art parallel imaging strategies. The experiments correspond to an 8-fold accelerated case with random noise of $\sigma=0.01$ added
    in k-space. The numbers in the caption show the PSNR values in
    dB. The $A^Hb$ had PSNR value of $24.33$~dB. 
    }
  \label{fig:comparison8f}
\end{figure*}

Figures~\ref{fig:comparison6f} and \ref{fig:comparison8f} visually
compare two different slices at 4x and 8x accelerated data acquisition
in the presence of Gaussian noise of $\sigma=0.01$.  The
testing slices are from a subject, whose data was not used for
training.
 It is evident from the zoomed portions that the
reconstruction quality by the proposed method is better than the
techniques compared against.

\begin{table}
\centering
\caption{Five-fold cross-validation results. The PSNR values are shown for 6x and 10x acceleration factors in the format: $\mean\;\pm\;std,\; \min/\max$. Column~1 represents the $i^{th}$ testing subject as Sub~$i$.}
\label{tab:crossValidation}
\begin{tabularx}{\linewidth}{@{\hspace{.2\tabcolsep}}ccc} \toprule
Testing on   & 6 fold acceleration      & 10 fold acceleration                 \\ \midrule
Sub~1 & $39.57\pm1.55,\;36.40/43.79 $& $37.95\pm1.68,\;34.30/42.48 $\\
Sub~2 & $39.56\pm1.23,\;35.13/43.85 $& $37.57\pm1.26,\;34.92/42.71 $\\
Sub~3 & $39.19\pm0.98,\;36.46/43.03 $& $37.38\pm1.15,\;33.58/41.44 $\\
Sub~4 & $37.73\pm1.15,\;34.07/41.11 $& $36.27\pm1.13,\;33.24/39.66 $\\
Sub~5 & $39.35\pm1.27,\;35.33/42.59 $& $37.59\pm1.16,\;33.77/40.73 $\\
Avg.  & $39.08\pm1.23,\;35.47/42.87 $& $37.35\pm1.27,\;33.96/41.40 $\\ \bottomrule
\end{tabularx}
\end{table}

Table~\ref{tab:crossValidation} shows the five-fold cross-validation results. In the cross-validation process, we did training with the data of four subjects and testing on the fifth subject. We repeated this process five times with different split of the training and testing sets. The average PSNR values obtained on each of the five testing subjects during cross-validation at 6x and 10x acceleration factors are summarized in Table~\ref{tab:crossValidation}. The average cross-validation results suggest that the proposed MoDL is robust enough to be used in practice. The average results on subject~4 are relatively less as compared to the other four subjects since this particular subject was of the significantly different physical structure. The results can be improved by adapting the data augmentation that considers different zoom factors during the training process.

 Table~\ref{tab:timing} reports the training and testing time for different algorithms compared against. The CSTV does not require any training and it was run on the CPU with parallel processing turned on. The deep learning techniques were run on the GPU.

\begin{table}
  \centering
  \caption{Training and testing times for different algorithms. The training time is mentioned in hours whereas the testing time is mentioned in seconds.  The testing time is mentioned for reconstructing all the 164 slices.}
  \label{tab:timing}
\begin{tabular}{cccccc} \toprule
              & CSTV    & SD-ET-WS & CG-ET-NS & CG-PD-NS & MoDL      \\ \midrule
training & --      & 3.6      & 10.8      & 7.6      & 10.6 \\
testing  & 162     & 8        & 28       & 49       & 28  \\ \bottomrule
\end{tabular}
\end{table}

\section{Discussion}
The main focus of this paper is to introduce the MoDL framework for general inverse problems and demonstrate its benefits over other deep learning strategies. In our experiments, we restricted the model to ten iterations and five-layer CNN due to GPU memory limitations. The performance of this preliminary implementation may be improved using strategies such as more training data, data augmentation, regularization priors, and drop-outs. Similarly, deeper CNNs with improved performance may be learned when more training data is available; our experiments show that when more training data is available, increasing the complexity of the network $\mathcal N_{\mathbf w}$ and repeating it as in Fig.~\ref{subfig:mainarch} is a better strategy than using different networks at different layers. However, these enhancements are beyond the scope of this work and will be dealt elsewhere.

We relied on a training strategy involving several variable density
sampling patterns to reduce the sensitivity of the algorithm to the
specific acquisition setting. However, this approach might have
resulted in the framework relying more on the lower k-space samples,
with a slight loss in high-frequency details. Improved results may be
possibly obtained by fixing the sampling pattern during training, as
pursued by most of the deep learning image recovery strategies~\cite{unser2017,barauniukDeepInverse2017,Zhang2017plugplaycnn,schlemper2017cascadeRueckert,jongCT}, albeit with increased sensitivity to acquisition settings.
We used a simple alternating  strategy to solve \eqref{maineqn} and hence unroll the network. Alternate approaches such as ADMM~\cite{Boyd2010} or the use of momentum terms~\cite{amir2009fistaSJIS} may have offered faster convergence. This will be a focus of our future work. 

We have not
  theoretically analyzed the convergence of the network in this
  work. Note that we use the proposed approach with a finite fixed number of iterations, where convergence issues are not too important. We will evaluate the benefit of using more iterations, as well as a detailed theoretical analysis of the framework in our future work.

\section{Conclusions}
We have introduced a model-based approach for image reconstruction
using a deep learned prior. The proposed MoDL framework combines the
power of data-driven learning with the physics derived model-based
framework. This framework provides a systematic approach for designing deep architectures for inverse
problems with the arbitrary structure. The introduction of the
optimization algorithm (CG) within a network layer allows extending
the model to complex forward models such as multi-channel MRI, where
the analytical inverse of the normal operator does not exist. Also,
this strategy offers the easy inclusion of additional image priors,
when available.

The sharing of the weights across iterations
facilitates the decoupling of the convergence from the complexity of
the network. The ability of the network to perform more iterations
without increasing the degrees of freedom reduces the risk of
overfitting, especially in medical imaging settings where training
data is scarce.
Our results show that the model provides improved results compared to state-of-the-art, despite the relatively smaller number of trainable parameters. Since we presented different sampling patterns during training, we obtained reduced sensitivity to acquisition parameters such as under-sampling ratio and amount of noise, which eliminates the need for training multiple large networks for each acquisition setting.

\bibliographystyle{IEEEtran}
\IEEEtriggeratref{25}


\begin{thebibliography}{10}
\providecommand{\url}[1]{#1}
\csname url@samestyle\endcsname
\providecommand{\newblock}{\relax}
\providecommand{\bibinfo}[2]{#2}
\providecommand{\BIBentrySTDinterwordspacing}{\spaceskip=0pt\relax}
\providecommand{\BIBentryALTinterwordstretchfactor}{4}
\providecommand{\BIBentryALTinterwordspacing}{\spaceskip=\fontdimen2\font plus
\BIBentryALTinterwordstretchfactor\fontdimen3\font minus
  \fontdimen4\font\relax}
\providecommand{\BIBforeignlanguage}[2]{{%
\expandafter\ifx\csname l@#1\endcsname\relax
\typeout{** WARNING: IEEEtran.bst: No hyphenation pattern has been}%
\typeout{** loaded for the language `#1'. Using the pattern for}%
\typeout{** the default language instead.}%
\else
\language=\csname l@#1\endcsname
\fi
#2}}
\providecommand{\BIBdecl}{\relax}
\BIBdecl

\bibitem{fessler2010magazine}
J.~A. Fessler, ``{Model-Based Image Reconstruction for MRI},'' \emph{{IEEE}
  Signal Process. Mag.}, vol.~27, no.~4, pp. 81--89, 2010.

\bibitem{fessler2002CT}
I.~A. Elbakri and J.~A. Fessler, ``Statistical image reconstruction for
  polyenergetic x-ray computed tomography,'' \emph{{IEEE} Trans. Med. Imag.},
  vol.~21, no.~2, pp. 89--99, 2002.

\bibitem{petUnser}
J.~Verhaeghe, D.~Van De~Ville, I.~Khalidov, Y.~D'Asseler, I.~Lemahieu, and
  M.~Unser, ``Dynamic {PET} reconstruction using wavelet regularization with
  adapted basis functions,'' \emph{{IEEE} Trans. Med. Imag.}, vol.~27, no.~7,
  pp. 943--959, July 2008.

\bibitem{unsreMicroscopyModelBased}
F.~Aguet, D.~Van De~Ville, and M.~Unser, ``Model-based {2.5-D} deconvolution
  for extended depth of field in brightfield microscopy,'' \emph{{IEEE} Trans.
  Image Process.}, vol.~17, no.~7, pp. 1144--1153, July 2008.

\bibitem{shiqianma2008}
S.~Ma, W.~Yin, Y.~Zhang, and A.~Chakraborty, ``{An Efficient Algorithm for
  Compressed MR Imaging using Total Variation and Wavelets},'' in
  \emph{Computer Vision and Pattern Recognition}, 2008, pp. 1--8.

\bibitem{Lingala2011}
S.~G. Lingala, Y.~Hu, E.~DiBella, and M.~Jacob, ``{Accelerated dynamic MRI
  exploiting sparsity and low-rank structure: kt SLR},'' \emph{{IEEE} Trans.
  Med. Imag.}, vol.~30, no.~5, pp. 1042--1054, 2011.

\bibitem{lingala2012blind}
S.~G. Lingala and M.~Jacob, ``A blind compressive sensing frame work for
  accelerated dynamic mri,'' in \emph{IEEE Int.~Symp.~Bio.~Imag.}\hskip 1em
  plus 0.5em minus 0.4em\relax IEEE, 2012, pp. 1060--1063.

\bibitem{ravishankar2015tsp}
S.~Ravishankar and Y.~Bresler, ``{L0 Sparsifying Transform Learning With
  Efficient Optimal Updates and Convergence Guarantees},'' \emph{{IEEE} Trans.
  Signal Process.}, vol.~63, no.~9, pp. 2389--2404, 2015.

\bibitem{bcs2013MathewsJacob}
S.~G. Lingala and M.~Jacob, ``{Blind Compressive Sensing Dynamic MRI},''
  \emph{{IEEE} Trans. Med. Imag.}, vol.~32, no.~6, pp. 1132--1145, 2013.

\bibitem{ronneberger2015unet}
O.~Ronneberger, P.~Fischer, and T.~Brox, ``U-net: Convolutional networks for
  biomedical image segmentation,'' in \emph{International Conference on Medical
  Image Computing and Computer-Assisted Intervention (MICCAI)}.\hskip 1em plus
  0.5em minus 0.4em\relax Springer, 2015, pp. 234--241.

\bibitem{heCVPR2016residualLearning}
K.~He, X.~Zhang, S.~Ren, and J.~Sun, ``{Deep Residual Learning for Image
  Recognition},'' in \emph{IEEE Conference on Computer Vision and Pattern
  Recognition}, 2016, pp. 770--778.

\bibitem{wangCTtmi2017}
H.~Chen, Y.~Zhang, M.~K. Kalra, F.~Lin, Y.~Chen, P.~Liao, J.~Zhou, and G.~Wang,
  ``{Low-Dose CT with a Residual Encoder-Decoder Convolutional Neural
  Network},'' \emph{{IEEE} Trans. Med. Imag.}, vol.~36, no.~12, pp. 2524--2535,
  2017.

\bibitem{wangDeepImagingIEEEaccess}
G.~Wang, ``{A Perspective on Deep Imaging},'' \emph{IEEE Access}, vol.~4,
  no.~nn, pp. 8914--8924, 2016.

\bibitem{zhang2017dncnn}
K.~Zhang, W.~Zuo, S.~Member, Y.~Chen, D.~Meng, and L.~Zhang, ``{Beyond a
  Gaussian Denoiser: Residual Learning of Deep CNN for Image Denoising},''
  \emph{{IEEE} Trans. Image Process.}, vol.~26, no.~7, pp. 3142--3155, 2017.

\bibitem{zhang2017magazine}
L.~Zhang and W.~Zuo, ``{Image Restoration: From Sparse and Low-Rank Priors to
  Deep Priors},'' \emph{{IEEE} Signal Process. Mag.}, vol.~34, no.~5, pp.
  172--179, 2017.

\bibitem{Zhang2017plugplaycnn}
K.~Zhang, W.~Zuo, S.~Gu, and L.~Zhang, ``{Learning Deep CNN Denoiser Prior for
  Image Restoration},'' in \emph{IEEE Conference on Computer Vision and Pattern
  Recognition}, 2017, pp. 2808--2817.

\bibitem{oneNetwork2017iccv}
J.~H.~R. Chang, C.-L. Li, B.~Poczos, B.~V. K.~V. Kumar, and A.~C.
  Sankaranarayanan, ``{One Network to Solve Them All --- Solving Linear Inverse
  Problems using Deep Projection Models},'' in \emph{IEEE International
  Conference on er Vision}, 2017, pp. 1--12.

\bibitem{schlemper2017cascadeRueckert}
J.~Schlemper, J.~Caballero, J.~V. Hajnal, A.~Price, and D.~Rueckert, ``{A Deep
  Cascade of Convolutional Neural Networks for MR Image Reconstruction},'' in
  \emph{Information Processing in Medical Imaging}, 2017, pp. 647--658.

\bibitem{hammernik}
K.~Hammernik, T.~Klatzer, E.~Kobler, M.~P. Recht, D.~K. Sodickson, T.~Pock, and
  F.~Knoll, ``{Learning a Variational Network for Reconstruction of Accelerated
  MRI Data},'' \emph{Magnetic resonance in Medicine}, vol.~79, no.~6, pp.
  3055--3071, 2017.

\bibitem{Mardani2018cvpr}
M.~Mardani, H.~Monajemi, V.~Papyan, S.~Vasanawala, D.~Donoho, and J.~Pauly,
  ``{Recurrent Generative Adversarial Networks for Proximal Learning and
  Automated Compressive Image Recovery},'' in \emph{IEEE Conference on Computer
  Vision and Pattern Recognition}, 2018, p.~na.

\bibitem{putzky}
P.~Putzky and M.~Willing, ``{Recurrent Inference Machines for Solving Inverse
  Problems},'' in \emph{arXiv}, 2017, pp. 1--12.

\bibitem{lista}
K.~Gregor and Y.~LeCun, ``Learning fast approximations of sparse coding,'' in
  \emph{International Conference on Machine Learning}.\hskip 1em plus 0.5em
  minus 0.4em\relax Omnipress, 2010, pp. 399--406.

\bibitem{storm}
S.~Poddar and M.~Jacob, ``Dynamic mri using smoothness regularization on
  manifolds (storm),'' \emph{{IEEE} Trans. Med. Imag.}, vol.~35, no.~4, pp.
  1106--1115, 2016.

\bibitem{modlStormICASSP2018}
S.~Biswas, H.~K. Aggarwal, S.~Poddar, and M.~Jacob, ``{ Model-based
  free-breathing cardiac MRI reconstruction using deep learned and STORM
  priors: MoDL-STORM},'' in \emph{IEEE International Conference on Acoustics,
  Speech, and Signal Processing}, 2018, p.~NA.

\bibitem{ongie2015super}
G.~Ongie and M.~Jacob, ``Super-resolution mri using finite rate of innovation
  curves,'' in \emph{IEEE Int.~Symp.~Bio.~Imag.}\hskip 1em plus 0.5em minus
  0.4em\relax IEEE, 2015, pp. 1248--1251.

\bibitem{gregtsp2017}
G.~Ongie, S.~Biswas, and M.~Jacob, ``{Convex recovery of continuous domain
  piecewise constant images from non-uniform Fourier samples},'' \emph{{IEEE}
  Trans. Signal Process.}, vol.~66, no.~1, pp. 236--250, 2017.

\bibitem{ista2003wavelet}
M.~A.~T. Figueiredo, R.~D. Nowak, S.~Member, and R.~D. Nowak, ``{An EM
  Algorithm for Wavelet-Based Image Restoration},'' \emph{{IEEE} Trans. Image
  Process.}, vol.~12, no.~8, pp. 906--916, 2003.

\bibitem{hdtv2012jacob}
Y.~Hu and M.~Jacob, ``{Higher Degree Total Variation (HDTV ) Regularization for
  Image Recovery},'' \emph{{IEEE} Trans. Image Process.}, vol.~21, no.~5, pp.
  2259--2271, 2012.

\bibitem{gregSIAM2016}
G.~Ongie and M.~Jacob, ``{Off-the-Grid Recovery of Piecewise Constant Images
  from Few Fourier Samples},'' \emph{SIAM on Imag. Sci.}, vol.~9, no.~3, pp.
  1004----1041, 2016.

\bibitem{alohaLee2016}
D.~Lee, K.~H. Jin, E.~Y. Kim, S.-H. Park, and J.~C. Ye, ``Acceleration of mr
  parameter mapping using annihilating filter-based low rank hankel matrix
  (aloha),'' \emph{Magnetic resonance in medicine}, vol.~76, no.~6, pp.
  1848--1864, 2016.

\bibitem{Haldar2014loraks}
J.~P. Haldar, ``{Low-Rank Modeling of Local k-Space Neighborhoods (LORAKS) for
  Constrained MRI},'' \emph{{IEEE} Trans. Med. Imag.}, vol.~33, no.~3, pp.
  668--681, 2014.

\bibitem{bm3d}
K.~Dabov, A.~Foi, V.~Katkovnik, and K.~Egiazarian, ``Image denoising by sparse
  3-d transform-domain collaborative filtering,'' \emph{{IEEE} Trans. Image
  Process.}, vol.~16, no.~8, pp. 2080--2095, 2007.

\bibitem{barauniukDeepInverse2017}
A.~Mousavi and R.~G. Baraniuk, ``{Learning to Invert: Signal Recovery via Deep
  Convolutional Networks},'' in \emph{IEEE Intel.~Conf.Acou.,~Speech, Sig
  Proces}, 2017, pp. 2272--2276.

\bibitem{unser2017}
K.~H. Jin, M.~T. McCann, E.~Froustey, and M.~Unser, ``{Deep Convolutional
  Neural Network for Inverse Problems in Imaging},'' \emph{{IEEE} Trans. Image
  Process.}, vol.~29, pp. 4509--4522, 2017.

\bibitem{Schlemper2017ruckertDynamic}
J.~Schlemper, J.~Caballero, J.~V. Hajnal, A.~N. Price, and D.~Rueckert, ``A
  deep cascade of convolutional neural networks for dynamic mr image
  reconstruction,'' \emph{{IEEE} Trans. Med. Imag.}, vol.~37, no.~2, pp.
  491--503, 2018.

\bibitem{Diamond2017}
S.~Diamond, V.~Sitzmann, F.~Heide, and G.~Wetzstein, ``{Unrolled Optimization
  with Deep Priors},'' in \emph{arXiv:1705.08041}, 2017, pp. 1--11.

\bibitem{chan2017plug}
S.~H. Chan, X.~Wang, and O.~A. Elgendy, ``{Plug-and-Play ADMM for Image
  Restoration: Fixed Point Convergence and Applications},'' \emph{IEEE
  Transactions on Computational Imaging}, vol.~3, no.~1, pp. 84--98, 2017.

\bibitem{admmnet}
y.~yang, J.~Sun, H.~Li, and Z.~Xu, ``Deep admm-net for compressive sensing
  mri,'' in \emph{Advances in Neural Information Processing Systems 29}, 2016,
  pp. 10--18.

\bibitem{numerical_recipes_c}
W.~H. Press, S.~A. Teukolsky, W.~T. Vetterling, and B.~P. Flannery, ``Numerical
  recipes in c,'' \emph{Cambridge University Press}, vol.~1, p.~3, 1988.

\bibitem{Kingma2015}
D.~P. Kingma and J.~L. Ba, ``{Adam: a Method for Stochastic Optimization},''
  \emph{International Conference on Learning Representations 2015}, pp. 1--15,
  2015.

\bibitem{Ioffe2015BN}
\BIBentryALTinterwordspacing
S.~Ioffe and C.~Szegedy, ``{Batch Normalization: Accelerating Deep Network
  Training by Reducing Internal Covariate Shift},'' in \emph{IEEE International
  Conference on Machine Learning}, 2015, pp. 448--456. [Online]. Available:
  \url{http://arxiv.org/abs/1502.03167}
\BIBentrySTDinterwordspacing

\bibitem{espirit2014}
M.~Uecker, P.~Lai, M.~J. Murphy, P.~Virtue, M.~Elad, J.~M. Pauly, S.~S.
  Vasanawala, and M.~Lustig, ``{ESPIRiT - An eigenvalue approach to
  autocalibrating parallel MRI: Where SENSE meets GRAPPA},'' \emph{Magnetic
  Resonance in Medicine}, vol.~71, no.~3, pp. 990--1001, 2014.

\bibitem{jongCT}
E.~Kang, J.~Min, and J.~C. Ye, ``A deep convolutional neural network using
  directional wavelets for low-dose x-ray ct reconstruction,'' \emph{Medical
  Physics}, vol.~44, no.~10, pp. e360--e375, 2017.

\bibitem{Boyd2010}
S.~Boyd, N.~Parikh, E.~Chu, B.~Peleato, and J.~Eckstein, ``{Distributed
  Optimization and Statistical Learning via the Alternating Direction Method of
  Multipliers},'' \emph{Foundations and Trends{\textregistered} in Machine
  Learning}, vol.~3, no.~1, pp. 1--122, 2010.

\bibitem{amir2009fistaSJIS}
A.~Beck and M.~Teboulle, ``{A Fast Iterative Shrinkage-Thresholding Algorithm
  for Linear Inverse Problems},'' \emph{SIAM J.~Imag.~Sci.}, vol.~2, no.~1, pp.
  183--202, 2009.

\end{thebibliography}

\end{document}